\documentclass[10pt,twocolumn,letterpaper]{article}

\usepackage[pagenumbers]{cvpr} 

\newcommand{\OURS}{DNF}
\usepackage{graphicx}
\usepackage{amsmath}
\usepackage{amssymb}
\usepackage{booktabs}
\usepackage{multirow}
\usepackage{diagbox}
\usepackage{comment}

\definecolor{cvprblue}{rgb}{0.21,0.49,0.74}
\usepackage[pagebackref,breaklinks,colorlinks,allcolors=cvprblue]{hyperref}

\def\paperID{16830} 
\def\confName{CVPR}
\def\confYear{2025}

\title{DNF: Unconditional 4D Generation with Dictionary-based Neural Fields}

\author{
{Xinyi Zhang\textsuperscript{1}\quad 
Naiqi Li\textsuperscript{2}\quad
Angela Dai\textsuperscript{1}} \\
\vspace{-0.3cm}
\\
\vspace{-0.3cm}
{
$^1$ Technical University of Munich\quad 
$^2$ Tsinghua University}\\
\\
\vspace{-0.3cm}
{
\href{https://xzhang-t.github.io/project/DNF}{https://xzhang-t.github.io/project/DNF}
} \\
}

\begin{document}

\begin{figure}
\twocolumn[{%
\renewcommand\twocolumn[1][]{#1}%
\maketitle
\begin{center}
    \centering
    \vspace{-0.2cm}
    \includegraphics[width=0.98\textwidth]{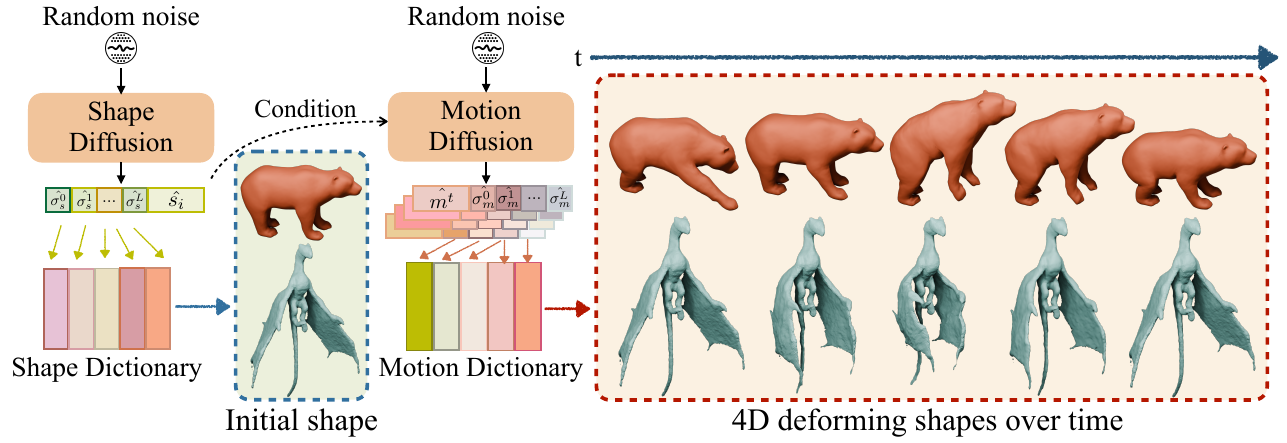}
    \vspace{-0.2cm}
    \caption{
    We propose \OURS{}, a dictionary-based representation for the unconditional generation of 4D deforming shapes, with a transformer-based diffusion model. Our method is capable of generating motions with superior shape quality and temporal consistency.
    }
    \label{figure:teaser}
\end{center}%
}]
\end{figure}

\begin{abstract}
While remarkable success has been achieved through diffusion-based 3D generative models for shapes, 4D generative modeling remains challenging due to the complexity of object deformations over time. 
We propose \OURS{}, a new 4D representation for unconditional generative modeling that efficiently models deformable shapes with disentangled shape and motion while capturing high-fidelity details in the deforming objects. 
To achieve this, we propose a dictionary learning approach to disentangle 4D motion from shape as neural fields.
Both shape and motion are represented as learned latent spaces, where each deformable shape is represented by its shape and motion global latent codes, shape-specific coefficient vectors, and shared dictionary information. 
This captures both shape-specific detail and global shared information in the learned dictionary. 
Our dictionary-based representation well balances fidelity, contiguity and compression -- combined with a transformer-based diffusion model, our method is able to generate effective, high-fidelity 4D animations. 
\end{abstract}

\section{Introduction}

3D shape representations have seen significant research in computer vision and graphics, with notable recent generative developments in neural field based representations ~\cite{mescheder2019occupancy,park2019deepsdf,mildenhall2021nerf}, which enable efficient capturing of high-fidelity detail in such a high-dimensional setting.
However, realistic generation requires not only 3D generation, but 4D - as the world is dynamic, encompassing and requiring motion to enable interactions, across wide-ranging applications such as content creation, mixed reality, simulation, and robotics.

Traditionally, template-based parametric models have been used to represent category-specific deforming objects, such as bodies\cite{loper2023smpl,osman2020star}, faces \cite{blanz2023morphable}, and hands \cite{MANO:SIGGRAPHASIA:2017,romero2022embodied}, where a fixed template mesh can be employed.
Advances in coordinate-MLP reprsentations to represent neural implicit fields have enabled a compact representation encompassing high-fidelity detail, with the ability arbitrarily query the coordinate-MLP for high resolutions.
Such coordinate-MLPs can be optimized to fit single shapes independently, reconstructing very high detail but lacking any shared structure across multiple shapes due to single-shape optimization.
These coordinate-MLP neural fields can also be learned across multiple objects, each represented by a latent code, which captures shared global structures but easily loses high-fidelity detail of individial shape details.
Additionally, 4D deforming shapes encompasses complexity in both shape and motion, and are more efficiently represented with disentangled shape (which remains the same over a deforming sequence) and the motion of that shape.

We thus propose \OURS{}, a new dictionary-learning based 4D representation that compactly represents deforming shapes while maintaining high fidelity; our representation enables unconditional 4D generation through diffusion generative modeling.
We introduce dictionary learning into a neural field representation. 
Inspired by template-based \cite{blanz2023morphable,loper2023smpl} and neural \cite{palafox2021npms,tang2022neural} parametric models, we first learn a neural field representation for both shape and motion of deforming 4D objects. 
We learn coarse latent shape and motion fields: the coarse latent shape space is learned for object shapes in their initial frame poses, and the motion field is conditioned on the shape latent feature to produce temporal flow from the initial frame to its following deformations. 

However, learning a single global latent-based representation for shape and motion tends to suffer from loss of detail, both in shape and temporal evolution. 
Thus, we construct a shared dictionary based on these representations, by decomposing the learned shape and motion MLPs using a singular value decomposition (SVD) \cite{golub1971singular}, and using the singular vector matrices as a shared dictionary.
We then fine-tune the singular values on each object; since the singular values can be viewed as the coefficient values of the linear combination of different elements in the dictionary, they are continuous and enable interpolation. 
Furthermore, to reduce the redundancy and improve the representation capabilities of the dictionary, we compress the dictionary by dropping the singular vectors with small singular values and then expand the dictionary with residual learning in row-rank form. 
This enables learning a powerful dictionary during the fine-tuning process, capable of representing 4D animations in a disentangled, compact fashion. 
Our dictionary-based representation characterizes 4D data in the form of latent and coefficient vectors per shape, accompanied with a shared dictionary, which effectively balances quality, contiguity and compression. 

We then train a transformer-based diffusion model on this representation, enabling unconditional generation of high-fidelity sequences. Due to our flexible dictionary, motions can also be generated for a given shape of a category not seen during training. With the diffusion out-painting, our generations can also be extended to a longer sequence with plausible motion.
Experiments on the DeformingThings4D \cite{li20214dcomplete} dataset demonstrate the effectiveness of our approach. The main contributions of our work are summarized as follows:
\begin{itemize}
\item{
We propose a novel, dictionary-based representation for 4D deforming shapes. A deforming shape is characterized by both shape-specific encodings (shape and motion latent, along with fine-tuned singular value coefficient vectors), along with a shared global dictionary, yield its 4D neural field representation.
} 

\item {
We construct a dictionary by decomposing globally-optimized shape and motion MLPs through singular value decomposition to enable a compact representation for fine-tuning shape-specific shape and motion parameters for high-fidelity 4D representations.
}

\item {
Our dictionary-based representation enables effective unconditional 4D generation by employing a transformer-based diffusion model on the learned dictionary representations, achieving state-of-the-art generation of deforming objects.
}
\end{itemize}

\begin{figure*}[tp]
\begin{center}
   \includegraphics[width=0.8\textwidth]{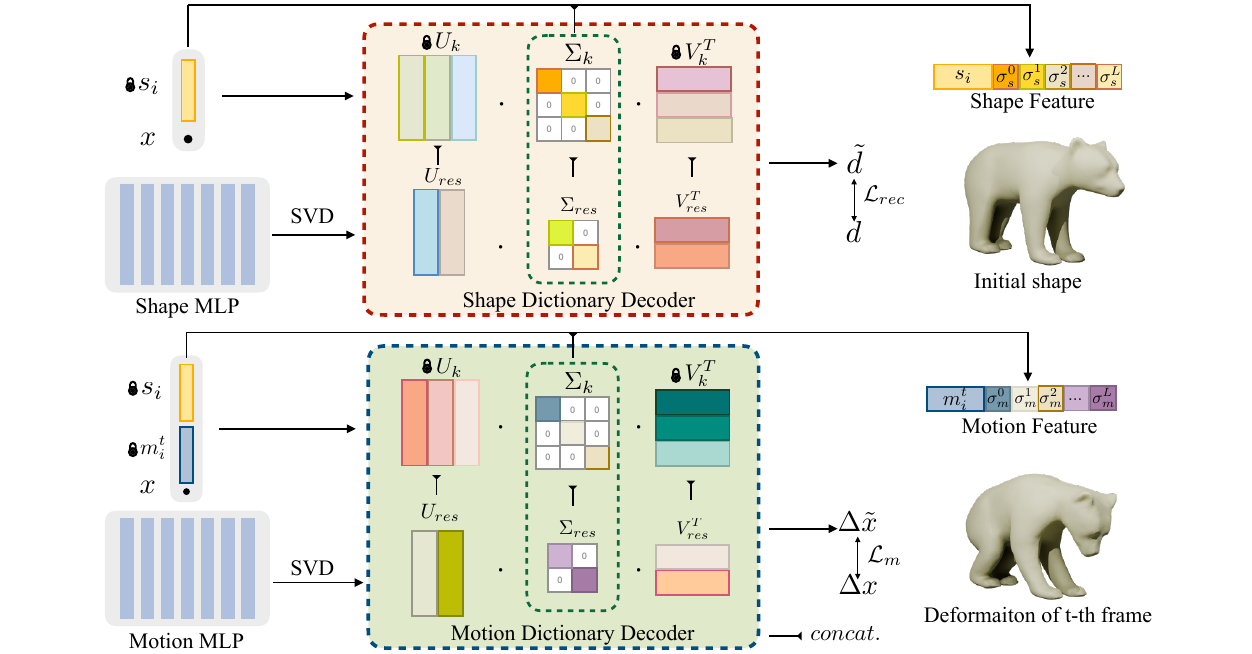}
\end{center}
    \vspace{-0.5cm}
    \caption{
    Overview for learning our 4D dynamic \OURS{} representation. 
    We first pre-train disentangled shape and motion MLPs with per-instance latents. 
    We then decompose the pre-trained MLPs using SVD to conduct dictionary-based fine-tuning of the singular values for each train instance, in order to more expressively capture local object detail. 
    We then obtain for each train instance its latent shape and motion codes as well as coefficient vectors, along with a globally shared dictionary. This effectively balances quality, contiguity and compression in the learned representation space. 
    }
\label{fig:rep}
\end{figure*}

\section{Related Work}
\paragraph{Representing 4D Deformable Shapes}
Inspired by the success of various advances in 3D representations for expressing static 3D objects, various approaches have been proposed for representing 4D deformable shapes.
For domain-specific modeling, such as for human bodies, heads, or hands, template-based parametric models have become widely used \cite{blanz2023morphable,loper2023smpl,MANO:SIGGRAPHASIA:2017,FLAME:SiggraphAsia2017}.
While a fixed template enables robust representation for specific category types, this limits shape expressivity and does not capture general deforming shapes across various categories.
Various methods have also recently been introduced to extend mesh-based parametric models to neural formulations \cite{tevet2023human,zhang2022motiondiffuse,kirschstein2024diffusionavatars, ao2023gesturediffuclip,tang2024dphms,palafox2021npms,palafox2022spams}, while continuing to leverage domain-specific knowledge, thereby constraining the approaches to domain-specific settings.

Recently, coordinate-field based MLP representations of neural fields have enabled a more flexible representation, capable of representing objects with arbitrary topologies and high resolution. 
For instance, Occupancy Flow~\cite{niemeyer2019occupancy} leverages an occupancy field based representation,  incorporating Neural-ODE~\cite{teschl2024ordinary} to simulate the velocity field for motion. LPDC \cite{tang2021learning} replaces the Neural-ODE with an MLP and learns local spatio-temporal codes, representing both shape and deformations.
NPMs~\cite{palafox2021npms} disentangles the shape and pose into separate latent spaces via two MLP networks, using a shape latent representing an SDF of the shape geometry and a pose latent representing the flow field from the canonical shape. While these methods show strong potential in the neural field representation, it remains challenging to accurately capture complex 4D dynamics, especially in non-rigid objects when using either ODE solvers or a single global latent vector coupled with an MLP network. 
In contrast, our method constructs a dictionary to encompass both global MLP-based optimization to fit general coarse shapes, along with per-shape specific fine-tuned parameters to achieve high-fidelity detail for each deforming shape sequence. 
This maintains a compact representation while enabling effective generative modeling through diffusion.

\paragraph{3D/4D Diffusion Models}
Generative models have achieved great success in generating new, similar and high-quality data by learning the underlying distributions of given data. 
In particulary, denoising diffusion probabilistic models~\cite{ho2020denoising, song2020denoising} have seen remarkable success in 2D \cite{rombach2022high,dhariwal2021diffusion,ho2022video,yang2024cogvideox} and even 3D generative modeling \cite{poole2022dreamfusion,vahdat2022lion,chou2022diffusionsdf,shue20233d,zhou20213d,tang2024diffuscene,roessle2024l3dg,meng2024lt3sd}, offering both stability of training as well as effective generation quality.


Various methods have been proposed to leverage diffusion modeling for 3D shape generation, using points or voxels \cite{nichol2022point,zhou20213d,chu2024diffcomplete}, as well as latent diffusion \cite{vahdat2022lion,chou2022diffusionsdf,cheng2023sdfusion} for more expressive modeling. 
In contrast, HyperDiffusion~\cite{erkocc2023hyperdiffusion} introduced generative modeling paradigm for encoding 3D or 4D shapes as their single shape optimized MLP weights of a neural implicit field, using diffusion to model the weight space of optimized MLPs through a diffusion process. 
However, due to the single shape optimization, the MLP weight space lacks strong shared structure between MLP weight encodings of different shapes, which hampers the generation process.
Motion2VecSets~\cite{cao2024motion2vecsets} further introduces a 4D neural representation employing latent vector sets describing shape and deformation flow, and uses a latent diffusion model for dynamic surface reconstruction from point cloud observations. 
We also propose to disentangle shape and motion in our 4D representation, but leverage a dictionary-based neural field learning to preserve quality while maintaining a compact, efficient representation with shared structure for unconditional diffusion modeling.

\section{Method}
\begin{figure*}[t]
\begin{center}
   \includegraphics[width=0.99\textwidth]{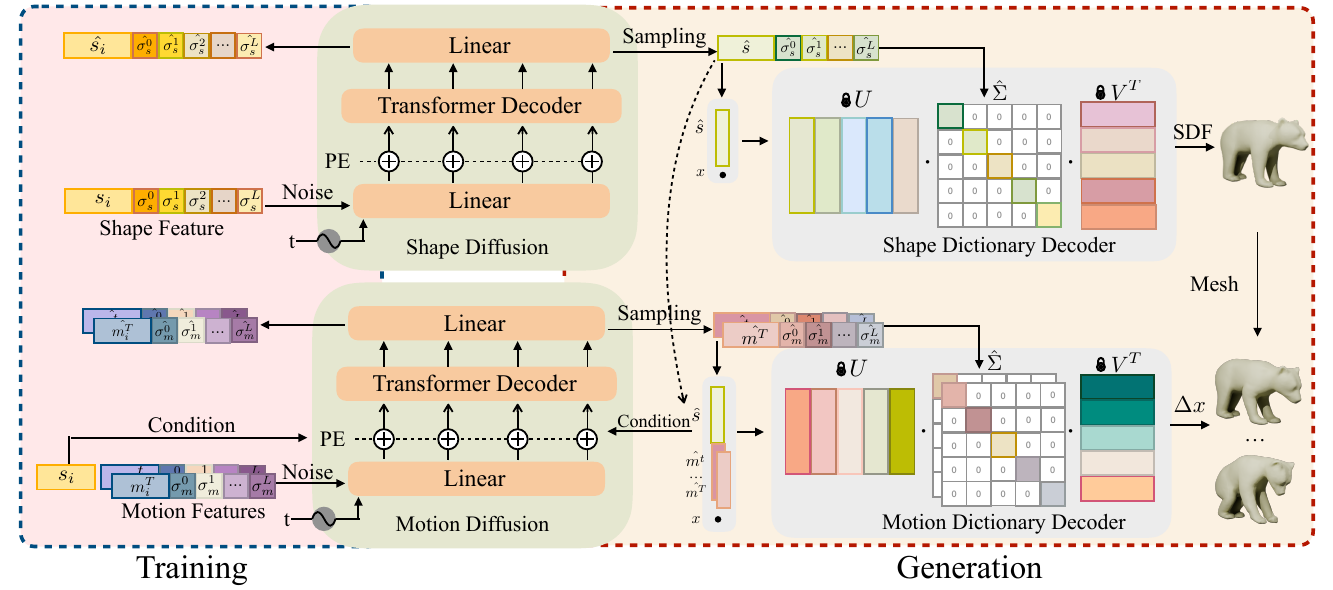}
\end{center}
\vspace{-0.7cm}
    \caption{
    Training and generation of our \OURS{}s for unconditional 4D synthesis.
    We employ transformer-based diffusion models to model the $\boldsymbol{\sigma}$ that modulate the shape and motion MLPs, along with shape and motion codes. At inference time, new samples can then be decoded to shape and motion to form a 4D deforming sequence.
    }
\label{fig:diff}
\end{figure*}
We introduce our  dictionary-learning based 4D shape representation, which allows a deforming object to be represented in the form of a shared dictionary, along with a shape-specific latent vector and coefficient vectors. 
We then use this representation for unconditional 4D generation, employing a diffusion model on our dictionary-based neural fields to generate new, high-fidelity 4D deforming shape sequences.

\subsection{Dictionary-based 4D Neural Fields}
To handle the challenges faced by 4D representations, we propose a novel neural field representation to balance shape fidelity, representation contiguity and compression.
This representation is learned from a training set comprising $S$ 4D sequences, containing $M$ deforming shape identities.
Inspired by Neural Parametric Models~\cite{palafox2021npms}, we first decompose a 4D sequence of a deforming shape into its shape and motion, using MLP-based coordinate fields. Each canonically-posed shape identity $i$ in the training set is encoded in a $D_s$-dimensional latent shape code $s_i$ through an auto-decoder.
Note that we do not assume that shapes are given in canonical poses, and simply use the initial shape in a train sequence as the canonical shape.
The shape MLP $f_{\Theta_s}$ predicts the implicit SDF $\tilde{d}$ for shape identities based on the shape code $s_i$ assigned to each $i$-th identity in $S$ shapes:
\begin{equation}
    f_{\Theta_s}(s_i,x) = \tilde{d}
\end{equation}

Based on the learned shape space, we then train a motion MLP conditioned on both the identity’s latent shape code and the corresponding $D_m$-dimensional latent motion code $m_i^t$ to predict a flow vector $\Delta \tilde{x}$ for the motion of the $t$-th frame of $i$-th identity:
\begin{equation}
f_{\Theta_m}(s_i,m_i^t,x) = \Delta \tilde{x}
\end{equation}

This produces coarse representations of shape and motion, but tends to lack detail when trained across diverse objects.
We thus also fine-tune these MLPs to better fit to individual 4D sequences while maintaining shared structure across different train elements.

\paragraph{Dictionary-based fine-tuning.}
To improve the representation power of the shape and motion MLPs, we introduce dictionary learning into their MLP fine-tuning.
We perform a singular value decomposition (SVD) on the MLP parameters, freezing the singular vectors and only fine-tuning the singular values. To reduce redundancy while further improving the representation power, we compress the dictionary (removing small singular values), and then extend it with low-rank residual learning. This representation learning is shown in Figure~\ref{fig:rep}.

More specifically, to improve the representation power of the shape and motion MLPs $f_{\Theta_s}$ and $f_{\Theta_m}$, we then keep the corresponding optimized shape and motion latents $\boldsymbol{s}$ and $\boldsymbol{m}$ fixed while fine-tuning the MLP parameters $\Theta_s$  and $\Theta_m$ for each 3D shape and deformation.
For simplicity of notation, we formulate the following for general MLP weights $\Theta$ and apply the same process for both $\Theta_s$  and $\Theta_m$.
Note that if we directly fine-tune the whole MLP parameters on each shape to  obtain shape-specific MLP weights, there would be no consistency in their weight space, resulting in poor continuity of the underlying representation for generative modeling. Thus, prior to fine-tuning, given an MLP network with layers $\ell=1,...,L$, and weights $\Theta$ = $\{W_{\ell} \in \mathbb{R}^{J\times F} \}_{\ell =1}^L$, we perform a layer-wise singular value decomposition (SVD) on the MLP parameters:
\begin{equation}
    W_{\ell} = U_{\ell}\Sigma_{\ell} V^T_{\ell},
\end{equation}
where $U_{\ell}\in \mathbb{R}^{J \times J}$ and $V_{\ell} \in \mathbb{R}^{F \times F}$ are matrices of singular vectors and $\Sigma_{\ell} \in \mathbb{R}^{J \times F} = \textrm{diag}(\boldsymbol{\sigma}_{\ell})$ is a diagonal matrix with descending non-negative singular values on its diagonal. 


The singular value decomposition can be written as 
\begin{equation}
W_{\ell} = \sum^r_{i=1} \sigma_{\ell,i} (\boldsymbol{u}_{\ell,i} \boldsymbol{v}^T_{\ell,i}),
\end{equation}
where $r \le \min(J,F)$ is the rank of $W_{\ell}$, $\boldsymbol{u}_{\ell,i}$ and $\boldsymbol{v}_{\ell,i}$ are the $i$-th column of $U_{\ell}$ and $V_{\ell}$, singular vectors of $W_{\ell}$.

Each  weight layer $W_{\ell}$ of the MLP parameters $\Theta$ can be viewed as a linear combination of elements in a dictionary, where $\boldsymbol{\sigma} = \{ \boldsymbol{\sigma}_{\ell} \}_{\ell =1}^L$ is the coefficient vector and the products of singular vectors in $U = \{U_{\ell}\}_{\ell =1}^L $ and $V= \{V_{\ell}\}_{\ell =1}^L$ form the dictionary. We further let $\boldsymbol{\sigma} = e^{\boldsymbol{\gamma}}$ to make sure the non-negativity of the coefficient values. We use the singular vector matrices  $U$ and $V$ to form our dictionary decoder $f_d$ and instantiate a copy of $\boldsymbol{\sigma}$ for each sample as $\{\boldsymbol{\sigma}^i\}_{i=1}^N$. 

For shape fitting, we decompose $\Theta_s$ to form the shape dictionary decoder $f_d^s$, fix the dictionary in $f_d^s$ and only fine-tune $\{\boldsymbol{\sigma}_s^i\}_{i=1}^S$ for each shape with a reconstruction loss:
\begin{equation}
\mathcal{L}_{rec}(\tilde{d},d) = |\textrm{clamp}(\tilde{d},\delta)-\textrm{clamp}(d,\delta)|,
\end{equation}
where $\textrm{clamp}(x,\delta) = \min(\delta, \max(-\delta, x))$ uses the parameter $\delta$ to control the distance from the surface, focusing learning on regions nearby and enhancing surface detail.

$\Theta_m$ is decomposed analogously: we build a motion dictionary decoder $f_d^m$ and fine-tune $\{\boldsymbol{\sigma}_m^i\}_{i=1}^M$ for each frame with an $\ell_1$-loss $\mathcal{L}_m$ on the flow prediction:
\begin{equation}
\mathcal{L}_{m}= |\Delta \tilde{x}-\Delta x|.
\end{equation}

Since the global latent space is continuous in its nature, we attach a list of coefficient vectors which is also continuous to the latent vector, ensuring the contiguity of their weight space and enabling to generalize to new samples.

\paragraph{Dictionary compression and extension}
A dictionary decomposed from a pre-trained MLP is capable of representing most cases by a linear combination of existing elements, but cannot fully represent all fine-scale local details, particularly for more complex shapes or deformations. 
Furthermore, not all the elements in the dictionary play an important role, making the full dictionary relatively inefficient. 
Thus, rather than using the dictionary directly obtained from the SVD, we first compress to reduce redundancy, and then extend the dictionary to enable more expressivity of detail. 

We compute an approximation to the matrix $\Theta$ by simply using only the most important components. 
Due to the nature of SVD, the data in the matrices $U$, $\Sigma$ and $V$ are sorted by their contribution to the matrix $\Theta$. 
We can then directly take $U_k$, $V_k^T$ and $\Sigma_k$, corresponding to the first $k$ columns of $U$ and $V$ and the upper left ($k \times k$)-square of $\Sigma$, to obtain the approximation:
\begin{equation}
\Theta \approx \widetilde{\Theta} = \lbrace \sum^k_{i=1} \sigma_{\ell,i} (\boldsymbol{u}_{\ell,i}  \boldsymbol{v}^T_{\ell,i}) \rbrace_{\ell=1}^L .
\end{equation}

After removing the superfluous elements in the dictionary, we then extend the dictionary with new, more relevant elements learned as residual offsets from the network parameters to further improve its representation capabilities:
\begin{equation}
\Theta' = \widetilde{\Theta} + \Delta \Theta,
\end{equation}
where $\Delta \Theta$ can also be written in the same form of SVD with low-rank matrices:
\begin{equation}
\Delta \Theta = \lbrace U_{res}^\ell \Sigma_{res}^\ell (V^T)_{res}^\ell \rbrace_{\ell=1}^L,
\end{equation}
with $U_{res}^\ell\in \mathbb{R}^{J \times rk}$, $\boldsymbol{\sigma}^\ell_{res} \in \mathbb{R}^{rk}$ , $V^\ell_{res} \in \mathbb{R}^{F \times rk}$, and $rk \ll J,F$.
Assuming that the residual vector matrices $U_{res}$ and $V_{res}$ can also serve as dictionaries, we aim to use them as singular vector matrices; that is, they should form orthogonal bases. To this end, we add an additional orthogonalization loss when optimizing the residual matrices:
\begin{equation}
\mathcal{L}_{orth} = |U^T_{res}U_{res}-I|+|V^T_{res}V_{res}-I|.
\end{equation}
With this orthogonalization loss, dictionary elements minimize redundancy and enhance interpretability, stability, and computational efficiency, resulting in more distinct and generalizable feature representations.

During this fine-tuning, we freeze $U_{k}$ and $V^T_{k}$ and optimize $U_{res}$ and $V^T_{res}$ among all train objects, in addition to their individual coefficient vectors $\{\boldsymbol{\sigma}_s^i\}_{i=1}^S$ for shapes and $\{\boldsymbol{\sigma}_m^i\}_{i=1}^M$ for motion of subsequent frames. 

As a result, we can represent each shape or motion with its original $D_s$ or $D_m$-dimensional latent code, concatenating a $L$-length coefficient vector list $\{ \boldsymbol{\sigma}_\ell \in \mathbb{R}^{(k+rk)} \}_{\ell=1}^L$, which learned during fine-tuning. We denote them as $\theta_s$ for shape features and $\theta_m$ for motion features. 
This representation is designed to maintain quality, contiguity in the representation space, and support a compact encoding. 
We can then further use it for generative modeling to synthesize new, high-quality 4D motions.

\subsection{Weight-Space Diffusion}

We then learn a generative model on our dictionary-based 4D representation, leveraging diffusion modeling.

\paragraph{Shape Diffusion}
We then model the weight space of $\theta_s$ through a diffusion process. Using a transformer backbone, our representation which is a $(L+1)$-length latent vector list, combining the shape code $s_i$ and $L$ coefficient vectors $\{\boldsymbol{\sigma}_s^{i,\ell}\}_{\ell=1}^L$, naturally split into $L+1$ tokens.


During diffusion modeling, as shown in Figure~\ref{fig:diff}, we gradually add gaussian noise $t$ times to $\theta_s$. A linear projection is then applied to the noised vector and the sinusoidal embedding of $t$. Afterwards, the projections are summed up with the position encoding vector on each token position and fed into a transformer decoder.
The transformer decoder consists of multiple self-attention layers, and predicts the denoised tokens with the simple objective \cite{ho2020denoising}:
\begin{equation}
\mathcal{L}_{simple} = E_{\theta_s \sim q(\theta_s),t \sim [1,T]}[||\theta_s - \hat{\theta_s}||^2_2].
\end{equation}
Then the denoised tokens are passed through a final output projection to produce the predicted denoised vectors $\hat{\theta_s}$.

During inference, we can then sample new $\hat{\theta_s}$ from random noise. 
We then split $\hat{\theta_s}$ into a latent vector $\hat{s}$ and a list of coefficient vectors $\hat{\boldsymbol{\sigma}_s^\ell}$ for each MLP layer.
With the shape dictionary decoder $f_d^s$, we can obtain a neural field with the generated $\hat{\boldsymbol{\sigma_s}}$ to present the generated shape with the predicted SDF:
\begin{equation}
\begin{aligned}
f_d^s(\hat{s},x,\hat{\boldsymbol{\sigma_s}}) = \tilde{d}.
\end{aligned}
\end{equation}

\paragraph{Motion Diffusion}
In order to model the motion sequence of a deforming shape, we train a diffusion model on windowed sequences of length $t$.
That is, for a motion sequence with $T$ original frames, we randomly pick subsequences with $t$ frames for training. 
These subsequences are constructed by concatenating the $t$ motion features in an extra time dimension $\boldsymbol{\theta}^t_m = \{\theta_m^i\}_{i=a}^{a+t}$ , conditioning on the shape code of the canonical shape. To introduce the shape conditions, we use a conditional cross-attention in addition to the self-attention layers in the transformer decoder.

To further maintain the frame order and improve coherence in the time dimension, we add an extra temporal self-attention on the time dimension afterwards.
The temporal self-attention is performed among tokens from different frames, but with the same positions in the $\{\theta_m^t\}$ (e.g., motion codes for different frames). 

Trained on a random subsequences of the original motion sequence, our motion diffusion is capable of generating sequences longer than $t$ frames through diffusion out-painting with a sliding window. We first generate a $t$-frame sequence, using the last $k$ frames as the context, and let the diffusion model in-paint the following $(t-k)$ frames, and iteratively repeat this process. In practice, our diffusion model is trained to generate 6-frame motions and uses the last 2 frames as context to in-paint the subsequent 4 frames, thus extending the generated motion sequence.


\section{Experiments}

We evaluate our approach on unconditional 4D motion generation, and demonstrate its ability to generalize to synthesizing new motions for unseen animal species.


\subsection{Experimental Setup}
\paragraph{Datasets.}
We use the DeformingThings4D~\cite{li20214dcomplete} dataset to train our approach and all baselines. 
DeformingThings4D contains 38 different shape identities for a total of 1227 animations, divided into training (75\%), validation (5\%), and test (20\%) subsets. The test sets are divided into unseen motions and unseen shapes, including unseen species.
We use the first frame of each train sequence to represent shape identities for shape training.
Shape SDFs are computed by sampling 200,000 points around the object surface and uniformly in the unit sphere. 
For motion sequences, we sample the first 16 frames of each sequence and sample 200,000 corresponding points for each frame of the sequence. 

\paragraph{Implementation details.}
For our shape and motion MLPs, we use 384-dimensional latent codes along with an 8-layer 512-dim MLP and 8-layer 1024-dim MLP, respectively. 
For the SVD-based decomposition, we compress the shape dictionary length from 512 to 384 and add a 256-rank residual matrix to expand the dictionary. For the motion dictionary, we compress its length from 1024 to 768 and add a 512-rank residual matrix. Then we fine-tune the coefficient vectors and the residual matrix at the same time, using 1000 epochs for shapes and 400 epochs for motion. After representation learning, we use a list of nine vectors (the original latent code and eight coefficient vectors for eight MLP layer) to represent each object. The shape/motion diffusion model is trained on two NVIDIA RTX A6000 GPUs for one day, for 1000 epochs. 

\begin{table}[tp]
\begin{center}
\resizebox{.95\columnwidth}{!}{
\begin{tabular}{c c c c}
\hline
Method & MMD $\downarrow$ & COV(\%) $\uparrow$ & 1-NNA(\%) $\downarrow$ \\ \hline
HyperDiffusion~\cite{erkocc2023hyperdiffusion} & 16.0 & 45.9 & 63.5 \\ 
Motion2VecSets~\cite{cao2024motion2vecsets} & 18.7 & 48.1 & 68.2 \\ 
Ours & \textbf{15.3} & \textbf{54.1} & \textbf{58.2}  \\
\hline
\end{tabular}
}
\end{center}
\vspace{-0.5cm}
\caption{Quantitative comparisons for 4D unconditional generation of animation sequences.
Our \OURS{} enable higher-quality generation through its expressive dictionary-based 4D representation.}
\label{tab:quan}
\end{table}

\paragraph{Baselines.}
We compare our unconditional generation results with state-of-the-art methods HyperDiffusion\cite{erkocc2023hyperdiffusion} and Motion2VecSets\cite{cao2024motion2vecsets}. HyperDiffusion generates the weights of 4D neural occupancy fields directly through a weight-space diffusion model. Motion2VecSets proposes a diffusion model designed for 4D dynamic surface reconstruction from sparse point clouds. To enable Motion2VecSets to produce unconditional generation results, we train the model in an unconditional setting. All baselines are trained on the same train split of DeformingThings4D as our method.

\paragraph{Evaluation metrics.}
Following previous works ~\cite{erkocc2023hyperdiffusion,vahdat2022lion,zhou20213d}, we use three Chamfer-based evaluation metrics, 1) Minimum Matching Distance (MMD), 2) Coverage (COV), and 3) 1-Nearest-Neighbor Accuracy (1-NNA), to measure generation quality.
\begin{figure*}[tp]
\begin{center}
   \includegraphics[width=0.98\textwidth]{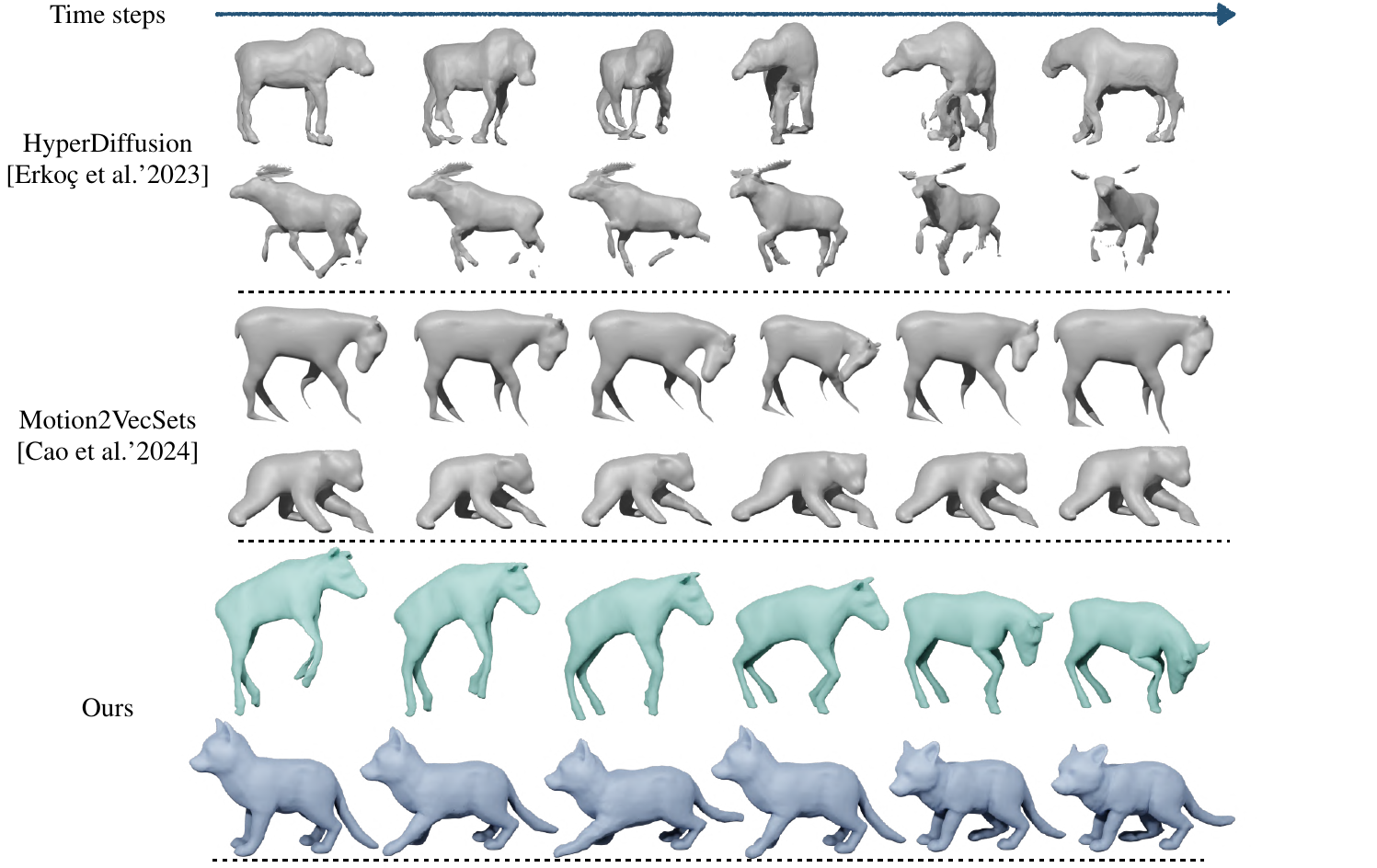}
\end{center}
\vspace{-0.5cm}
    \caption{
    Qualitative comparison with state of the art. Our dictionary-based approach enables generating 4D sequences with higher shape fidelity and temporal consistency.
    }
   
\label{fig:vis}
\end{figure*}

\subsection{Unconditional Motion Generation}
We compare with state of the art on unconditional 4D generation, generating  16-frame animal motion sequences.
For our method, we first generate 6 frames and then continually extend by 4 frames, using the last 2 frames as context.
As shown in Tab~\ref{tab:quan}, we achieve notably improved results compared to the baselines.
Fig.~\ref{fig:vis} shows a visual comparison, illustrating our improved visual quality and temporal consistency in the generated motions. 

In particular, HyperDiffusion suffers from poor shape quality with broken or missing legs during movements, due to the lack of continuity in the weight space of HyperDiffusion. 
Motion2VecSets preserves finer shape details, but struggles to generate coherent motions without conditional guidance. In contrast, our dictionary-based approach not only exhibits superior shape quality in each frame, but also demonstrates significantly improved temporal consistency throughout the entire motion sequences.

   

\begin{figure}[t]
\begin{center}
   \includegraphics[width=0.98\columnwidth]{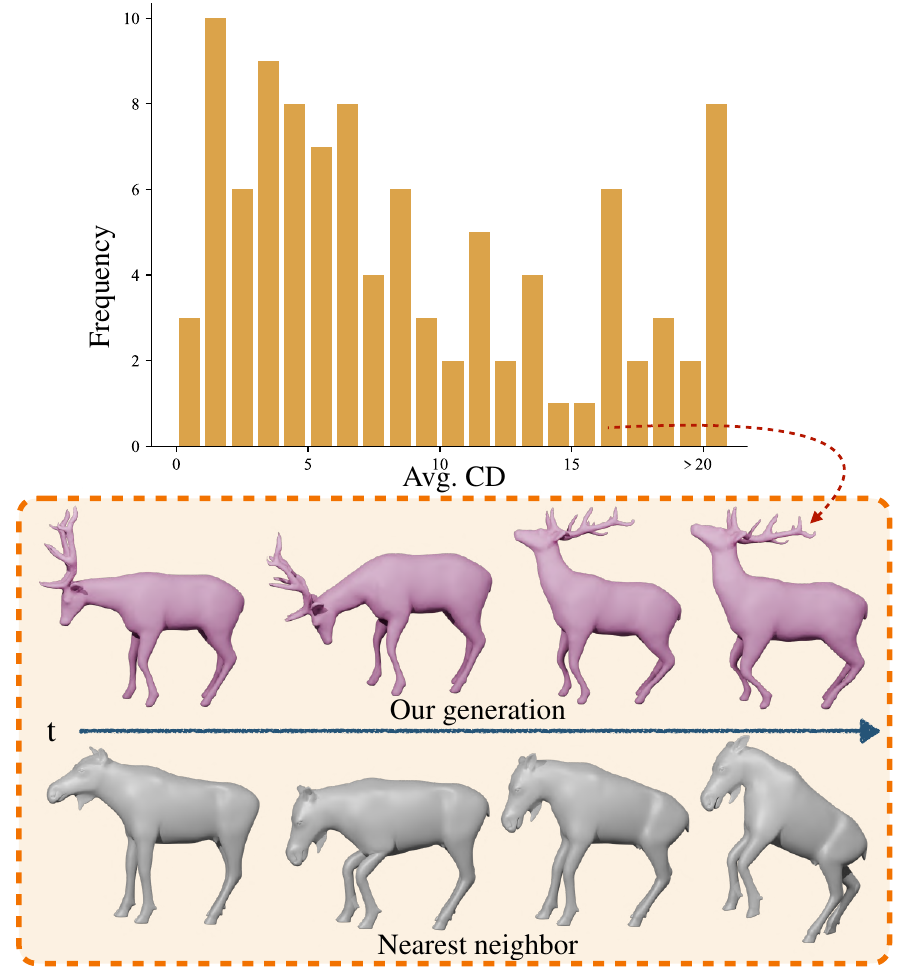}
\end{center}
\vspace{-0.5cm}
    \caption{
     Distribution of the average chamfer distance for all generations of our method to their nearest neighbors from the train set, showing that our method is able to synthesize new motions.
    }
\label{fig:novelty}
\end{figure}

\paragraph{Novelty analysis.}
We assess \OURS{}'s capability to generate novel 4D sequences.
We sample 100 random deforming shape sequences from our trained diffusion model, and retrieve their nearest neighbors in the training set by calculating the average Chamfer Distance between frames in each sequence. We plot the distribution of average Chamfer Distances for all generated motions in Fig.~\ref{fig:novelty}, and present a comparison between our generated motion and its closest counterpart in the training set.
Though the initial frames appear more similar, the motions diverge as they progress.

\subsection{Ablations}
To evaluate the effectiveness of our representation in capturing finer local details, we conduct ablation studies focusing on the impact of dictionary-based fine-tuning and the decoupling of shape and motion spaces.

\paragraph{Effect of the dictionary-based fine-tuning.}
To verify the effectiveness of our dictionary-based fine-tuning in capturing finer local details, we compare our method with NPMs in terms of shape reconstruction quality. As shown in Tab.~\ref{tab:ablation}, we compare the average Chamfer Distance between the ground truth meshes and the reconstructed meshes over the first 16 frames of each motion.
NPMs uses only a disentangled global latent representation for each deforming shape, which struggles to capture fine details, in contrast to our instance-specific compressed, weight-space fine-tuning.

\paragraph{Effect of decoupling shape and motion.}
Another approach to fitting deforming shapes is to directly fine-tune the shape code (with or without the coefficient vector list $\boldsymbol{\sigma_s}$) of the initial shape to accommodate subsequent deformations, while keeping the global MLP fixed (denoted as $\boldsymbol{s}_\textrm{ft}$ and $\boldsymbol{s}\_\boldsymbol{\sigma_s}_\textrm{ft}$). As shown in Tab.~\ref{tab:ablation}, when fine-tuning with $\boldsymbol{\sigma_s}$, the reconstruction performance surpasses NPMs', but remains inferior to our decoupled method. Additionally, decoupling shape and motion space not only improves the reconstruction quality, but also enables our diffusion model to generate motions for unseen animal species.

\begin{table}[t]
\begin{center}
\resizebox{.9\columnwidth}{!}{
\begin{tabular}{c c c c c}
\hline
Method & NPMs~\cite{palafox2021npms} & $\boldsymbol{s}_\textrm{ft}$ &$\boldsymbol{s}\_ \boldsymbol{\sigma_s}_\textrm{ft}$ & Ours \\ \hline
CD & 0.128 & 0.154 & 0.096 & \textbf{0.067} \\
\hline
\end{tabular}
}
\end{center}
\vspace{-0.5cm}
\caption{
Quantitative ablations for shape reconstruction: (i) NPMs uses only shape and pose MLPs without fine-tuning; (ii) fitting $\boldsymbol{s}$ of the first shape to the deformations; (iii) fitting both $\boldsymbol{s}$ and $\boldsymbol{\sigma_s}$ of the first shape to the deformations and (iv) ours. We compute the average Chamfer Distance for the reconstructions of training sequences with 16 frames (10,000 points, multiplied by $10^3$). Our \OURS{} demonstrates significantly improved representation capabilities.}

\label{tab:ablation}
\end{table}

\subsection{Generating Motions for Unseen Shape Species}
\begin{figure}[t]
\begin{center}
   \includegraphics[width=0.98\columnwidth]{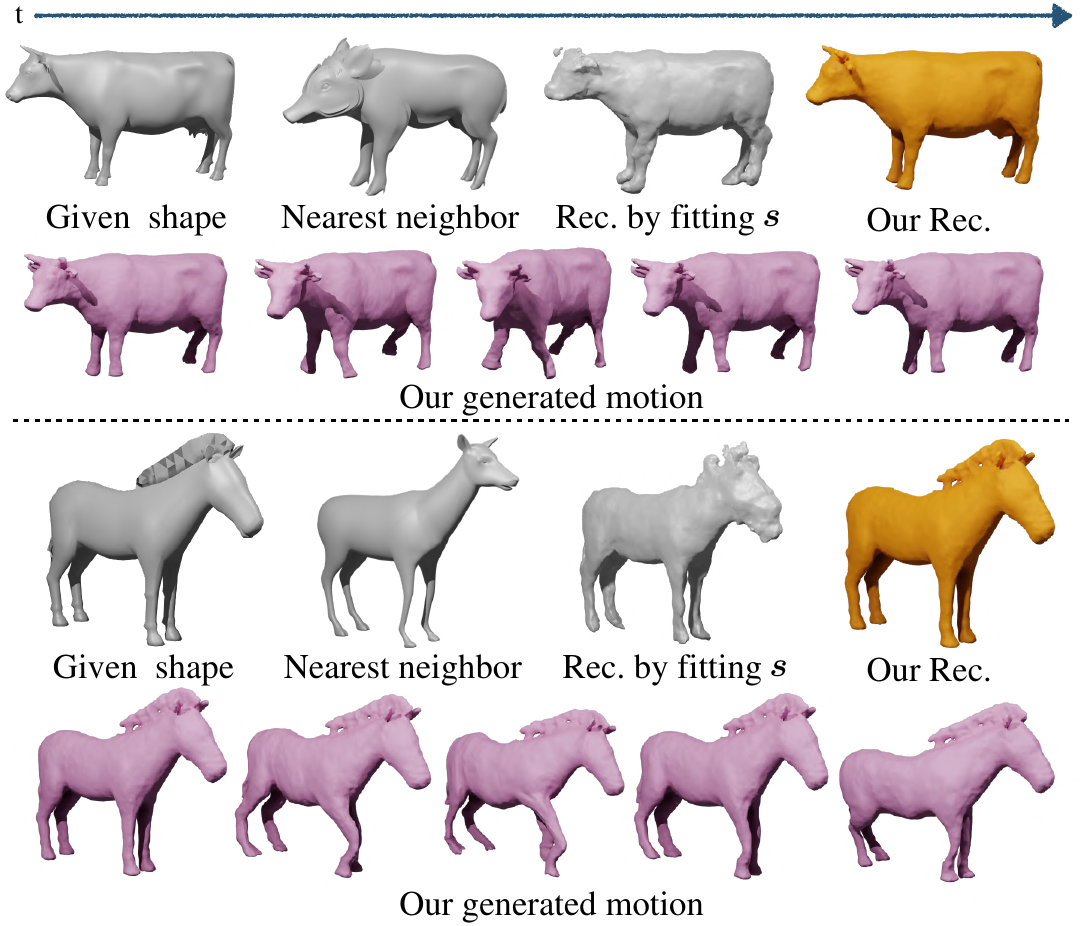}
\end{center}
\vspace{-0.5cm}
    \caption{
     Visualizations of shape fitting and motion generation on unseen animal species. Given the shape identity of an unseen species, which differs significantly from its nearest neighbor in the training set, and is difficult to fit to when optimizing only shape code $\boldsymbol{s}$, our method is capable of generating a high-quality reconstruction while producing plausible motions for the new shape.
    }
   
\label{fig:unseen}
\end{figure}

Our generative model can even generalize to generate plausible motions for unseen shape identities, including unseen species. 
Given a mesh of a new shape identity, we leverage our learned shape dictionary to obtain a neural field that represents the shape by optimizing a new shape code and coefficient vector list. As shown in Fig.~\ref{fig:unseen}, for animal species not seen during training, our method significantly improves reconstruction quality compared to using only the shape code for fitting, which captures the general shape but struggles to represent fine local details. Moreover, our motion diffusion model can also generalize to these new shape conditions, producing realistic global and local deformations for unseen animal species.

\vspace{-0.5cm}
\paragraph{Limitations.}
While our \OURS{} demonstrates potential for a more expressive, compact 4D representation space, various limitations remain. For instance, our learned spaces do not consider physical constraints, which can result in volume distortion or physically incorrect motions to be synthesized.
Additionally, our diffusion generative modeling operates only on per-instance specific encodings (individual latents and vector coefficients), which makes training compact and efficient, but leaves the modeling process unaware of the full dictionary decoding process and the final surface to be decoded.
\section{Conclusion}
We have presented a new, dictionary-based representation for 4D deforming objects that maintains a compact, contiguous latent representation to disentangle shape and motion for high-fidelity unconditional 4D generation. 
We leverage a weight-space representation of shape and motion for 4D objects, using compressed dictionary-based fine-tuning to maintain local detail across a diverse array of shapes.
This enables training a diffusion model on our dictionary-based representation to synthesize new deforming sequences.
We believe this will enable new opportunities in generative modeling for high-dimensional, complex data.


\paragraph{Acknowledgments.}
This project was supported by the ERC Starting Grant SpatialSem (101076253) and the TUM Georg Nemetschek Institute Artificial Intelligence for the Built World.

{
    \small
    \bibliographystyle{ieee_fullname}
    \bibliography{egbib}

\begin{thebibliography}{41}
\providecommand{\natexlab}[1]{#1}
\providecommand{\url}[1]{\texttt{#1}}
\expandafter\ifx\csname urlstyle\endcsname\relax
  \providecommand{\doi}[1]{doi: #1}\else
  \providecommand{\doi}{doi: \begingroup \urlstyle{rm}\Url}\fi

\bibitem[Ao et~al.(2023)Ao, Zhang, and Liu]{ao2023gesturediffuclip}
Tenglong Ao, Zeyi Zhang, and Libin Liu.
\newblock Gesturediffuclip: Gesture diffusion model with clip latents.
\newblock \emph{ACM Transactions on Graphics (TOG)}, 42\penalty0 (4):\penalty0 1--18, 2023.

\bibitem[Blanz and Vetter(2023)]{blanz2023morphable}
Volker Blanz and Thomas Vetter.
\newblock A morphable model for the synthesis of 3d faces.
\newblock In \emph{Seminal Graphics Papers: Pushing the Boundaries, Volume 2}, pages 157--164. 2023.

\bibitem[Cao et~al.(2024)Cao, Luo, Zhang, Nie{\ss}ner, and Tang]{cao2024motion2vecsets}
Wei Cao, Chang Luo, Biao Zhang, Matthias Nie{\ss}ner, and Jiapeng Tang.
\newblock Motion2vecsets: 4d latent vector set diffusion for non-rigid shape reconstruction and tracking.
\newblock In \emph{Proceedings of the IEEE/CVF Conference on Computer Vision and Pattern Recognition}, pages 20496--20506, 2024.

\bibitem[Cheng et~al.(2023)Cheng, Lee, Tuyakov, Schwing, and Gui]{cheng2023sdfusion}
Yen-Chi Cheng, Hsin-Ying Lee, Sergey Tuyakov, Alex Schwing, and Liangyan Gui.
\newblock {SDFusion}: Multimodal 3d shape completion, reconstruction, and generation.
\newblock In \emph{CVPR}, 2023.

\bibitem[Chou et~al.(2022)Chou, Bahat, and Heide]{chou2022diffusionsdf}
Gene Chou, Yuval Bahat, and Felix Heide.
\newblock Diffusionsdf: Conditional generative modeling of signed distance functions.
\newblock \emph{arXiv preprint arXiv:2211.13757}, 2022.

\bibitem[Chu et~al.(2023)Chu, Xie, Mo, Li, Nie{\ss}ner, Fu, and Jia]{chu2024diffcomplete}
Ruihang Chu, Enze Xie, Shentong Mo, Zhenguo Li, Matthias Nie{\ss}ner, Chi-Wing Fu, and Jiaya Jia.
\newblock Diffcomplete: Diffusion-based generative 3d shape completion.
\newblock \emph{Advances in Neural Information Processing Systems}, 2023.

\bibitem[Dhariwal and Nichol(2021)]{dhariwal2021diffusion}
Prafulla Dhariwal and Alexander Nichol.
\newblock Diffusion models beat gans on image synthesis.
\newblock \emph{Advances in neural information processing systems}, 34:\penalty0 8780--8794, 2021.

\bibitem[Erko{\c{c}} et~al.(2023)Erko{\c{c}}, Ma, Shan, Nie{\ss}ner, and Dai]{erkocc2023hyperdiffusion}
Ziya Erko{\c{c}}, Fangchang Ma, Qi Shan, Matthias Nie{\ss}ner, and Angela Dai.
\newblock Hyperdiffusion: Generating implicit neural fields with weight-space diffusion.
\newblock \emph{arXiv preprint arXiv:2303.17015}, 2023.

\bibitem[Golub and Reinsch(1971)]{golub1971singular}
Gene~H Golub and Christian Reinsch.
\newblock Singular value decomposition and least squares solutions.
\newblock In \emph{Handbook for Automatic Computation: Volume II: Linear Algebra}, pages 134--151. Springer, 1971.

\bibitem[Ho et~al.(2020)Ho, Jain, and Abbeel]{ho2020denoising}
Jonathan Ho, Ajay Jain, and Pieter Abbeel.
\newblock Denoising diffusion probabilistic models.
\newblock \emph{Advances in neural information processing systems}, 33:\penalty0 6840--6851, 2020.

\bibitem[Ho et~al.(2022)Ho, Salimans, Gritsenko, Chan, Norouzi, and Fleet]{ho2022video}
Jonathan Ho, Tim Salimans, Alexey Gritsenko, William Chan, Mohammad Norouzi, and David~J Fleet.
\newblock Video diffusion models.
\newblock \emph{Advances in Neural Information Processing Systems}, 35:\penalty0 8633--8646, 2022.

\bibitem[Kirschstein et~al.(2024)Kirschstein, Giebenhain, and Nie{\ss}ner]{kirschstein2024diffusionavatars}
Tobias Kirschstein, Simon Giebenhain, and Matthias Nie{\ss}ner.
\newblock Diffusionavatars: Deferred diffusion for high-fidelity 3d head avatars.
\newblock In \emph{Proceedings of the IEEE/CVF Conference on Computer Vision and Pattern Recognition}, pages 5481--5492, 2024.

\bibitem[Li et~al.(2017)Li, Bolkart, Black, Li, and Romero]{FLAME:SiggraphAsia2017}
Tianye Li, Timo Bolkart, Michael.~J. Black, Hao Li, and Javier Romero.
\newblock Learning a model of facial shape and expression from {4D} scans.
\newblock \emph{ACM Transactions on Graphics, (Proc. SIGGRAPH Asia)}, 36\penalty0 (6):\penalty0 194:1--194:17, 2017.

\bibitem[Li et~al.(2021)Li, Takehara, Taketomi, Zheng, and Nie{\ss}ner]{li20214dcomplete}
Yang Li, Hikari Takehara, Takafumi Taketomi, Bo Zheng, and Matthias Nie{\ss}ner.
\newblock 4dcomplete: Non-rigid motion estimation beyond the observable surface.
\newblock In \emph{Proceedings of the IEEE/CVF International Conference on Computer Vision}, pages 12706--12716, 2021.

\bibitem[Loper et~al.(2023)Loper, Mahmood, Romero, Pons-Moll, and Black]{loper2023smpl}
Matthew Loper, Naureen Mahmood, Javier Romero, Gerard Pons-Moll, and Michael~J Black.
\newblock Smpl: A skinned multi-person linear model.
\newblock In \emph{Seminal Graphics Papers: Pushing the Boundaries, Volume 2}, pages 851--866. 2023.

\bibitem[Meng et~al.(2024)Meng, Li, Nie{\ss}ner, and Dai]{meng2024lt3sd}
Quan Meng, Lei Li, Matthias Nie{\ss}ner, and Angela Dai.
\newblock Lt3sd: Latent trees for 3d scene diffusion.
\newblock \emph{arXiv preprint arXiv:2409.08215}, 2024.

\bibitem[Mescheder et~al.(2019)Mescheder, Oechsle, Niemeyer, Nowozin, and Geiger]{mescheder2019occupancy}
Lars Mescheder, Michael Oechsle, Michael Niemeyer, Sebastian Nowozin, and Andreas Geiger.
\newblock Occupancy networks: Learning 3d reconstruction in function space.
\newblock In \emph{Proceedings of the IEEE/CVF conference on computer vision and pattern recognition}, pages 4460--4470, 2019.

\bibitem[Mildenhall et~al.(2021)Mildenhall, Srinivasan, Tancik, Barron, Ramamoorthi, and Ng]{mildenhall2021nerf}
Ben Mildenhall, Pratul~P Srinivasan, Matthew Tancik, Jonathan~T Barron, Ravi Ramamoorthi, and Ren Ng.
\newblock Nerf: Representing scenes as neural radiance fields for view synthesis.
\newblock \emph{Communications of the ACM}, 65\penalty0 (1):\penalty0 99--106, 2021.

\bibitem[Nichol et~al.(2022)Nichol, Jun, Dhariwal, Mishkin, and Chen]{nichol2022point}
Alex Nichol, Heewoo Jun, Prafulla Dhariwal, Pamela Mishkin, and Mark Chen.
\newblock Point-e: A system for generating 3d point clouds from complex prompts.
\newblock \emph{arXiv preprint arXiv:2212.08751}, 2022.

\bibitem[Niemeyer et~al.(2019)Niemeyer, Mescheder, Oechsle, and Geiger]{niemeyer2019occupancy}
Michael Niemeyer, Lars Mescheder, Michael Oechsle, and Andreas Geiger.
\newblock Occupancy flow: 4d reconstruction by learning particle dynamics.
\newblock In \emph{Proceedings of the IEEE/CVF international conference on computer vision}, pages 5379--5389, 2019.

\bibitem[Osman et~al.(2020)Osman, Bolkart, and Black]{osman2020star}
Ahmed~AA Osman, Timo Bolkart, and Michael~J Black.
\newblock Star: Sparse trained articulated human body regressor.
\newblock In \emph{Computer Vision--ECCV 2020: 16th European Conference, Glasgow, UK, August 23--28, 2020, Proceedings, Part VI 16}, pages 598--613. Springer, 2020.

\bibitem[Palafox et~al.(2021)Palafox, Bo{\v{z}}i{\v{c}}, Thies, Nie{\ss}ner, and Dai]{palafox2021npms}
Pablo Palafox, Alja{\v{z}} Bo{\v{z}}i{\v{c}}, Justus Thies, Matthias Nie{\ss}ner, and Angela Dai.
\newblock Npms: Neural parametric models for 3d deformable shapes.
\newblock In \emph{Proceedings of the IEEE/CVF International Conference on Computer Vision}, pages 12695--12705, 2021.

\bibitem[Palafox et~al.(2022)Palafox, Sarafianos, Tung, and Dai]{palafox2022spams}
Pablo Palafox, Nikolaos Sarafianos, Tony Tung, and Angela Dai.
\newblock Spams: Structured implicit parametric models.
\newblock In \emph{Proceedings of the IEEE/CVF Conference on Computer Vision and Pattern Recognition}, pages 12851--12860, 2022.

\bibitem[Park et~al.(2019)Park, Florence, Straub, Newcombe, and Lovegrove]{park2019deepsdf}
Jeong~Joon Park, Peter Florence, Julian Straub, Richard Newcombe, and Steven Lovegrove.
\newblock Deepsdf: Learning continuous signed distance functions for shape representation.
\newblock In \emph{Proceedings of the IEEE/CVF conference on computer vision and pattern recognition}, pages 165--174, 2019.

\bibitem[Poole et~al.(2022)Poole, Jain, Barron, and Mildenhall]{poole2022dreamfusion}
Ben Poole, Ajay Jain, Jonathan~T Barron, and Ben Mildenhall.
\newblock Dreamfusion: Text-to-3d using 2d diffusion.
\newblock \emph{arXiv preprint arXiv:2209.14988}, 2022.

\bibitem[Roessle et~al.(2024)Roessle, M{\"u}ller, Porzi, Bul{\`o}, Kontschieder, Dai, and Nie{\ss}ner]{roessle2024l3dg}
Barbara Roessle, Norman M{\"u}ller, Lorenzo Porzi, Samuel~Rota Bul{\`o}, Peter Kontschieder, Angela Dai, and Matthias Nie{\ss}ner.
\newblock L3dg: Latent 3d gaussian diffusion.
\newblock In \emph{SIGGRAPH Asia 2024 Conference Papers}, 2024.

\bibitem[Rombach et~al.(2022)Rombach, Blattmann, Lorenz, Esser, and Ommer]{rombach2022high}
Robin Rombach, Andreas Blattmann, Dominik Lorenz, Patrick Esser, and Bj{\"o}rn Ommer.
\newblock High-resolution image synthesis with latent diffusion models.
\newblock In \emph{Proceedings of the IEEE/CVF conference on computer vision and pattern recognition}, pages 10684--10695, 2022.

\bibitem[Romero et~al.(2017)Romero, Tzionas, and Black]{MANO:SIGGRAPHASIA:2017}
Javier Romero, Dimitrios Tzionas, and Michael~J. Black.
\newblock Embodied hands: Modeling and capturing hands and bodies together.
\newblock \emph{ACM Transactions on Graphics, (Proc. SIGGRAPH Asia)}, 36\penalty0 (6), 2017.

\bibitem[Romero et~al.(2022)Romero, Tzionas, and Black]{romero2022embodied}
Javier Romero, Dimitrios Tzionas, and Michael~J Black.
\newblock Embodied hands: Modeling and capturing hands and bodies together.
\newblock \emph{arXiv preprint arXiv:2201.02610}, 2022.

\bibitem[Shue et~al.(2023)Shue, Chan, Po, Ankner, Wu, and Wetzstein]{shue20233d}
J~Ryan Shue, Eric~Ryan Chan, Ryan Po, Zachary Ankner, Jiajun Wu, and Gordon Wetzstein.
\newblock 3d neural field generation using triplane diffusion.
\newblock In \emph{Proceedings of the IEEE/CVF Conference on Computer Vision and Pattern Recognition}, pages 20875--20886, 2023.

\bibitem[Song et~al.(2020)Song, Meng, and Ermon]{song2020denoising}
Jiaming Song, Chenlin Meng, and Stefano Ermon.
\newblock Denoising diffusion implicit models.
\newblock \emph{arXiv preprint arXiv:2010.02502}, 2020.

\bibitem[Tang et~al.(2021)Tang, Xu, Jia, and Zhang]{tang2021learning}
Jiapeng Tang, Dan Xu, Kui Jia, and Lei Zhang.
\newblock Learning parallel dense correspondence from spatio-temporal descriptors for efficient and robust 4d reconstruction.
\newblock In \emph{Proceedings of the IEEE/CVF Conference on Computer Vision and Pattern Recognition}, pages 6022--6031, 2021.

\bibitem[Tang et~al.(2022)Tang, Markhasin, Wang, Thies, and Nie{\ss}ner]{tang2022neural}
Jiapeng Tang, Lev Markhasin, Bi Wang, Justus Thies, and Matthias Nie{\ss}ner.
\newblock Neural shape deformation priors.
\newblock \emph{Advances in Neural Information Processing Systems}, 35:\penalty0 17117--17132, 2022.

\bibitem[Tang et~al.(2024{\natexlab{a}})Tang, Dai, Nie, Markhasin, Thies, and Nie{\ss}ner]{tang2024dphms}
Jiapeng Tang, Angela Dai, Yinyu Nie, Lev Markhasin, Justus Thies, and Matthias Nie{\ss}ner.
\newblock Dphms: Diffusion parametric head models for depth-based tracking.
\newblock In \emph{Proceedings of the IEEE/CVF Conference on Computer Vision and Pattern Recognition}, pages 1111--1122, 2024{\natexlab{a}}.

\bibitem[Tang et~al.(2024{\natexlab{b}})Tang, Nie, Markhasin, Dai, Thies, and Nie{\ss}ner]{tang2024diffuscene}
Jiapeng Tang, Yinyu Nie, Lev Markhasin, Angela Dai, Justus Thies, and Matthias Nie{\ss}ner.
\newblock Diffuscene: Denoising diffusion models for generative indoor scene synthesis.
\newblock In \emph{Proceedings of the IEEE/CVF conference on computer vision and pattern recognition}, pages 20507--20518, 2024{\natexlab{b}}.

\bibitem[Teschl(2024)]{teschl2024ordinary}
Gerald Teschl.
\newblock \emph{Ordinary differential equations and dynamical systems}.
\newblock American Mathematical Society, 2024.

\bibitem[Tevet et~al.(2023)Tevet, Raab, Gordon, Shafir, Cohen-or, and Bermano]{tevet2023human}
Guy Tevet, Sigal Raab, Brian Gordon, Yoni Shafir, Daniel Cohen-or, and Amit~Haim Bermano.
\newblock Human motion diffusion model.
\newblock In \emph{The Eleventh International Conference on Learning Representations}, 2023.

\bibitem[Vahdat et~al.(2022)Vahdat, Williams, Gojcic, Litany, Fidler, Kreis, et~al.]{vahdat2022lion}
Arash Vahdat, Francis Williams, Zan Gojcic, Or Litany, Sanja Fidler, Karsten Kreis, et~al.
\newblock Lion: Latent point diffusion models for 3d shape generation.
\newblock \emph{Advances in Neural Information Processing Systems}, 35:\penalty0 10021--10039, 2022.

\bibitem[Yang et~al.(2024)Yang, Teng, Zheng, Ding, Huang, Xu, Yang, Hong, Zhang, Feng, et~al.]{yang2024cogvideox}
Zhuoyi Yang, Jiayan Teng, Wendi Zheng, Ming Ding, Shiyu Huang, Jiazheng Xu, Yuanming Yang, Wenyi Hong, Xiaohan Zhang, Guanyu Feng, et~al.
\newblock Cogvideox: Text-to-video diffusion models with an expert transformer.
\newblock \emph{arXiv preprint arXiv:2408.06072}, 2024.

\bibitem[Zhang et~al.(2022)Zhang, Cai, Pan, Hong, Guo, Yang, and Liu]{zhang2022motiondiffuse}
Mingyuan Zhang, Zhongang Cai, Liang Pan, Fangzhou Hong, Xinying Guo, Lei Yang, and Ziwei Liu.
\newblock Motiondiffuse: Text-driven human motion generation with diffusion model.
\newblock \emph{arXiv preprint arXiv:2208.15001}, 2022.

\bibitem[Zhou et~al.(2021)Zhou, Du, and Wu]{zhou20213d}
Linqi Zhou, Yilun Du, and Jiajun Wu.
\newblock 3d shape generation and completion through point-voxel diffusion.
\newblock In \emph{Proceedings of the IEEE/CVF international conference on computer vision}, pages 5826--5835, 2021.

\end{thebibliography}
}
\clearpage
\setcounter{page}{1}
\maketitlesupplementary

\begin{abstract}
In this supplementary file, we provide additional detail about our network architecture (Section~\ref{sec:netarch}), along with further elaboration on implementation details (Section~\ref{sec:implementationdetails}). We also refer to reader to the supplemental video for further qualitative results of our DNF for 4D synthesis. 
\end{abstract}

\section{Network Architecture Details}
\label{sec:netarch}
\subsection{Dictionary Decoder}
With a pre-trained shape and motion MLP, we first conduct SVD to each linear layer of the MLP and compress the matrices $U \in \mathbb{R}^{J \times J} $, $\Sigma \in \mathbb{R}^{J \times F}$ and $V \in \mathbb{R}^{F \times F}$ to $U_k \in \mathbb{R}^{J \times k}$,$V_k \in \mathbb{R}^{F \times k}$ and $\Sigma_k \in \mathbb{R}^{k \times k}$.  For each layer in the MLP, we then use two linear layers $N_U \in \mathbb{R}^{J \times k}$ and $N_V \in \mathbb{R}^{F \times k}$ to play the role as $U$ and $V$, replacing the original linear layer $N \in \mathbb{R}^{J \times F}$.
To extend the dictionary, we use another two linear layers $N_{U_r} \in \mathbb{R}^{J \times rk}$ and $N_{V_r} \in \mathbb{R}^{F \times rk}$ to learn the residual. During the fine-tuning, we freeze the parameters of $N_U$ and $N_V$ and optimize $N_{U_r}$, $N_{V_r}$ and $\boldsymbol{\sigma}$.

\subsection{Shape Diffusion}
For each shape feature, consisting of nine $(L+1)$ vectors (one original latent code and eight coefficient vectors corresponding to eight MLP layers), we naturally split them into nine tokens. Each token is projected to the same dimension, set to 1280 in our implementation. The projected tokens are then summed with positional encoding vectors corresponding to their positions and fed into a transformer decoder. The transformer decoder, composed of 32 self-attention layers, predicts the denoised tokens.

\subsection{Motion Diffusion}
The overall architecture of motion diffusion is similar to that of shape diffusion but operates on a sequence of motions with $t$ frames as input. The $t$ motion features are concatenated along an additional time dimension, and a positional encoding is added in this dimension to ensure the correct order of the generated motions. Similarly, we project these $(t \times L)$ tokens to an inner dimension and add positional encoding vectors based on their token positions. In the motion diffusion model, each layer of the transformer decoder contains three attention layers:  
\begin{enumerate}
    \item A \textbf{spatial self-attention layer} to aggregate tokens within each frame,  
    \item A \textbf{condition cross-attention layer} to incorporate shape conditions, and  
    \item A \textbf{temporal self-attention layer} to aggregate tokens from the same position across different frames (e.g., motion codes of different frames).
\end{enumerate}
In the sampling stage, our motion diffusion is capable of generating sequences longer than $t$ frames through diffusion out-painting with a sliding window. We first generate a $t$-frame sequence, using the last $k$ frames as the context, and let the diffusion model in-paint the following $(t-k)$ frames, and iteratively repeat this process. 

To be more specific, given the motion features \(\{\theta_m^k\}\) of the last \(k\) frames, we append \((t-k)\) vectors, \(\{\theta_m^{(t-k)}\}\), which are initialized as random noise of the same size as the motion features. The goal is to denoise \(\{\theta_m^{(t-k)}\}\) using the context provided by \(\{\theta_m^k\}\).

For each denoising time step \(d\), we aim to denoise \(\{\theta_m^{(t-k)}\}_d\) into \(\{\theta_m^{(t-k)}\}_{d-1}\). To achieve this, we first apply a \(d\)-step diffusion process to \(\{\theta_m^k\}\), obtaining a noised version, \(\{\theta_m^k\}_d\), which is then concatenated with \(\{\theta_m^{(t-k)}\}_d\). Subsequently, our motion diffusion model denoises the combined vectors, producing \(\{\theta_m^{(t-k)}\}_{d-1}\) using a DDIM sampler.

In practice, our diffusion model is trained to generate 6-frame motions and uses the last 2 frames as context to in-paint the subsequent 4 frames, thus extending the generated motion sequence.

\section{Implementation Details}
\label{sec:implementationdetails}
\subsection{Data processing}
\paragraph{Shape space.} For each shape identity in the train dataset, we sample 200k points on the given mesh. We then calculate its grid SDF with resolution equals to $256$, sampling 50k points uniformly within the unit bounding box and 150k random near-surface points within a distance of 0.02 from the surface of the shape.

\paragraph{Pose space.} Following previous work \cite{palafox2021npms}, we sample 200k surface points on each shape identity and store the barycentric weights for each sampled point at the same time. Each point is then randomly disturbed with a small noise $\mathcal{N}(0,\Sigma^2)$ along the normal direction of the corresponding triangle in the mesh, with $\Sigma \in \mathbb{R}^3$ a diagonal covariance matrix with entries $\Sigma_{ii}=\sigma$. Then, for each t-th deforming shape for the identity, we compute corresponding points by using the same barycentric weights and the noise to sample the deformed mesh. In our experiments, we sample 50\% surface points ($\sigma=0$) and 50\% with $\sigma=0.002$.

\subsection{Data augmentation}
When training the motion diffusion model, we apply data augmentation techniques to enhance the model's robustness. Specifically, for each motion subsequence, we reverse the frame order to create a new training sample, which significantly improves the continuity of the generated motions. 
Additionally, we distribute the shape condition \(\mathcal{S}\) using a few-step diffusion process, defined as 
\[
\mathcal{S}_t = f(\mathcal{S}_{t-1}, \epsilon_t), \quad \text{for } t = 1, \dots, T,
\]
where \(f\) represents the diffusion forward function, \(\epsilon_t\) is the added noise, and \(T\) is the total number of steps. Here, we choose \(T\) randomly from the range \([0, 50]\).


\end{document}